\documentclass[letterpaper, 10 pt, conference]{ieeeconf}  

\IEEEoverridecommandlockouts                              

\usepackage{amsmath}
\usepackage{amssymb}
\usepackage{booktabs}
\usepackage{multirow}
\usepackage{gensymb}
\usepackage{dsfont}
\let\labelindent\relax
\usepackage{enumitem}
\usepackage{pifont}
\usepackage{graphicx}
\usepackage{xcolor}
\usepackage[pagebackref,breaklinks,colorlinks]{hyperref}

\newcommand{\myparagraph}[1]{\textbf{#1 ---}}
\definecolor{RectangleGreen}{rgb}{0.478,0.576,0.521}
\definecolor{ObjectGreen}{rgb}{0.188,0.435,0.113}
\newcommand{\thinrule}{\specialrule{0.1pt}{0.2pt}{0.2pt}}

\title{\LARGE \bf
Teaching Agents how to Map:\\ Spatial Reasoning for Multi-Object Navigation
}

\author{Pierre Marza$^{1}$, Laetitia Matignon$^{2}$, Olivier Simonin$^{3}$ and Christian Wolf$^{4}$ \\ Project Page: \href{https://pierremarza.github.io/projects/teaching_agents_how_to_map/}{https://pierremarza.github.io/projects/teaching\_agents\_how\_to\_map/}
\thanks{$^{1}$Pierre Marza is with LIRIS, UMR CNRS 5205, Université de Lyon, INSA-Lyon, Villeurbanne, France. {\tt\small pierre.marza@insa-lyon.fr},}%
\thanks{$^{2}$Laetitia Matignon is with Univ Lyon, UCBL, CNRS, INSA-Lyon, LIRIS, UMR5205, F-69622 Villeurbanne, France. {\tt\small laetitia.matignon@univ-lyon1.fr}}%
\thanks{$^{3}$Olivier Simonin is with INSA Lyon, CITI Lab, INRIA Chroma team, Villeurbanne, France. {\tt\small olivier.simonin@insa-lyon.fr}}%
\thanks{$^{4}$Christian Wolf
is with Naver Labs Europe, France 
{\tt\small christian.wolf@naverlabs.com}
}
}

\begin{document}

\maketitle
\thispagestyle{empty}
\pagestyle{empty}

\begin{abstract}

In the context of visual navigation, the capacity to map a novel environment is necessary for an agent to exploit its observation history in the considered place and efficiently reach known goals. 
This ability can be associated with spatial reasoning, where an agent is able to perceive spatial relationships and regularities, and discover object characteristics. Recent work introduces learnable policies parametrized by deep neural networks and trained with Reinforcement Learning (RL). In classical RL setups, the capacity to map and reason spatially is learned end-to-end, from reward alone. In this setting, we introduce supplementary supervision in the form of auxiliary tasks designed to favor the emergence of spatial perception capabilities in agents trained for a goal-reaching downstream objective. We show that learning to estimate metrics quantifying the spatial relationships between an agent at a given location and a goal to reach has a high positive impact in Multi-Object Navigation settings. Our method significantly improves the performance of different baseline agents, that either build an explicit or implicit representation of the environment, even matching the performance of incomparable oracle agents taking ground-truth maps as input. A learning-based agent from the literature trained with the proposed auxiliary losses was the winning entry to the \textit{Multi-Object Navigation Challenge}, part of the \textit{CVPR 2021 Embodied AI Workshop}.
\end{abstract}

\section{INTRODUCTION}
\label{sec:intro}

\noindent
Navigating in a previously unseen environment requires different abilities, among which is mapping, i.e. the capacity to build a representation of the environment. The agent can then reason on this map and act efficiently towards its goal. How biological species map their environment is still an open area of research~\cite{peer2020structuring, warren2017wormholes}. In robotics, spatial representations have taken diverse forms, for instance metric maps~\cite{elfes1989using, bresson2017simultaneous} or topological maps~\cite{shatkay1997learning, thrun1998learning}. Most of these variants have lately been presented in neural counterparts, i.e. involving artificial neural networks --- metric neural maps~\cite{DBLP:conf/iclr/ParisottoS18, DBLP:conf/pkdd/BeechingD0020, Henriques_2018_CVPR} or neural topological maps~\cite{beeching2020learning, Chaplot_2020_CVPR} learned from RL or with supervision. 

This work focuses on improving the RL-based training strategy of autonomous agents parametrized by deep neural networks. We explore the question whether \textbf{the emergence of mapping and spatial reasoning capabilities can be favored by the use of spatial auxiliary tasks} that are related to a downstream objective. We target the problem of \textit{Multi-Object Navigation}~\cite{DBLP:conf/nips/WaniPJCS20}, where an agent must reach a sequence of specified objects in a particular order within a previously unknown environment. Such a task is interesting because it requires an agent to recall the position of previously encountered objects it will have to reach later in the sequence. 
This work does not introduce a new agent architecture, but rather showcases the impact of augmenting the vanilla RL training of state-of-the-art (SOTA) agents selected in~\cite{DBLP:conf/nips/WaniPJCS20} to solve the \textit{Multi-Object Navigation} task.
Augmenting the RL training of agents with auxiliary tasks has shown promise in many recent works introducing several variants~\cite{DBLP:conf/iclr/MirowskiPVSBBDG17, DBLP:conf/iclr/JaderbergMCSLSK17, lample2017playing, ye2021auxiliary, aux_pointgoal_custom}. These different formulations are presented in more details in section~\ref{sec:related}. Our work belongs to the group of supervised auxiliary tasks, with an application to 3D complex and photo-realistic environments, and  specifically targets the learning of mapping and spatial reasoning, which has not been the scope of previous work.

\begin{figure}[t]
    \centering
    \includegraphics[width=0.23\textwidth]{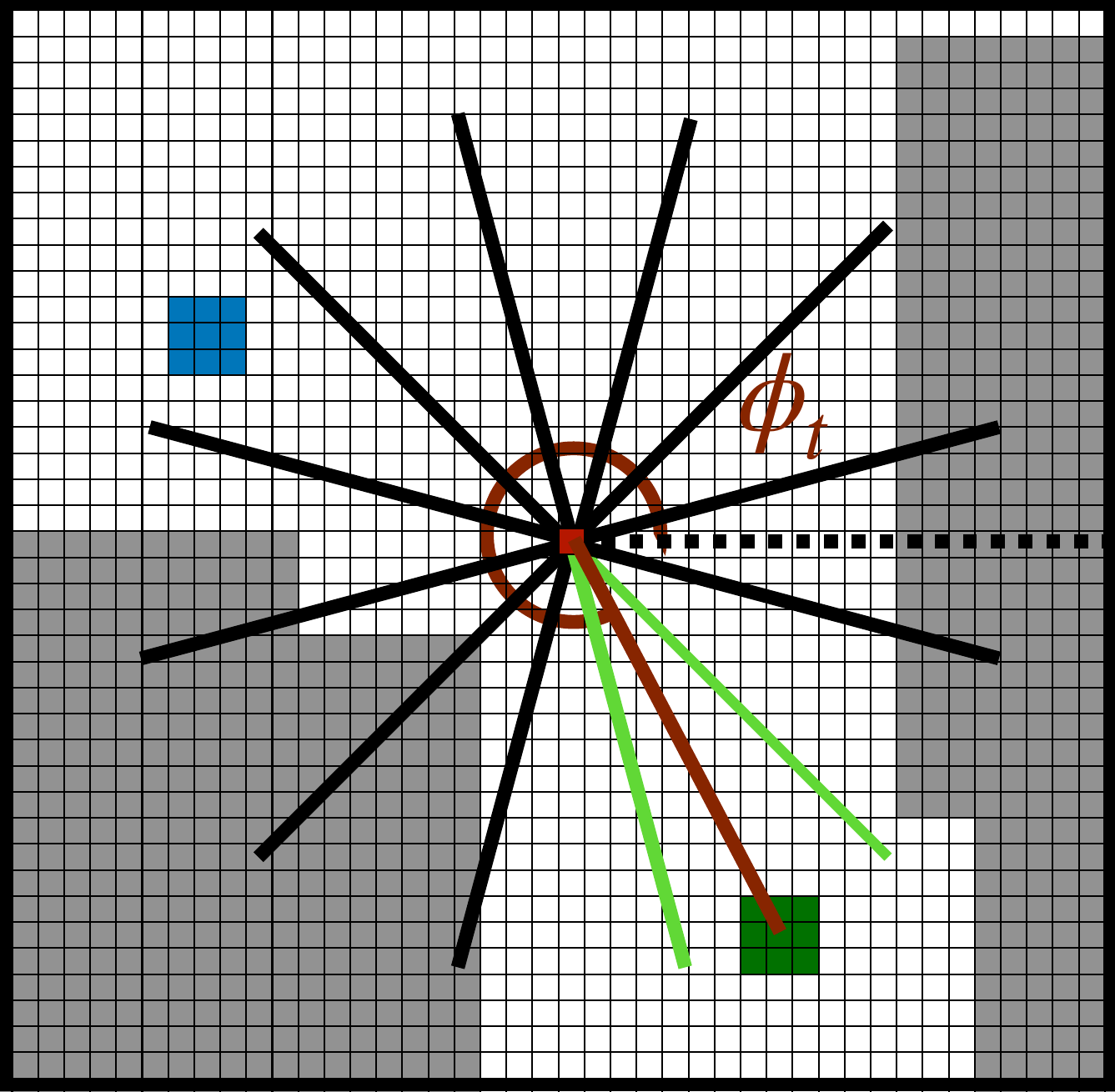}
    \includegraphics[width=0.23\textwidth]{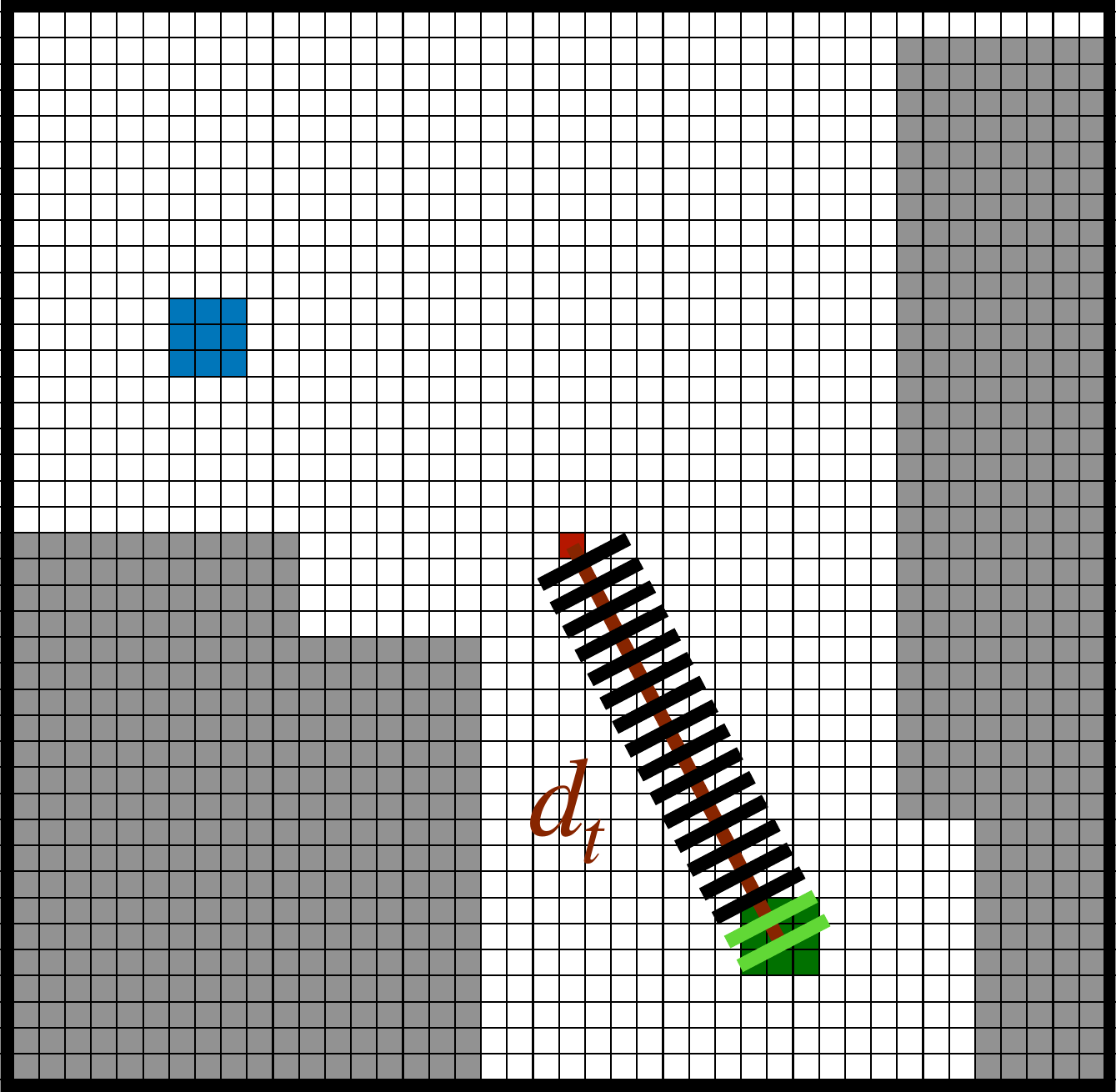}
    \caption{In the context of Deep-RL for Multi-Object Navigation, two auxiliary tasks predict the direction (\textit{left}) and the distance (\textit{right}) to the next object to retrieve \emph{if it has been observed during the episode}. The {\color{ObjectGreen} \textbf{green object}} (square) is the current target, and other targets exist and have already been found or might be required to be found later (ex: blue object / square). Red dot: position of the agent. White and grey cells, respectively, indicate free space and obstacles.  Both, the angle $\phi_t$ and the distance $d_t$ between the center of the map (i.e. the agent) and the target at time $t$ are discretized and associated with a class label. A third sub-task is to predict if the current target has already been within the agent's field of view during the episode.}
    \label{fig:aux_tasks}
\end{figure}

We take inspiration from behavioral studies of human spatial navigation~\cite{ekstrom2018human}. Experiments with human subjects aim at evaluating the spatial knowledge they acquire when navigating a given environment. In~\cite{ekstrom2018human}, two important measures are referred  as the \textit{sense of direction} and \textit{judgement of relative distance}. Regarding knowledge of direction, a well-known task is \textit{scene- and orientation- dependent pointing} (SOP), where participants must point to a specified location that is not currently within their field of view. Being able to assess its relative position compared to other objects in the world is critical to navigate properly, and disorientation is considered a main issue. In addition to direction, evaluating the distance to landmarks is also of high importance.

We conjecture that an agent able to estimate the location of target objects relative to its current pose will implicitly extract more useful representations of the environment and navigate more efficiently. A fundamental skill for such an agent is thus to remember previously encountered objects. Our auxiliary supervision targets exactly this ability. 
Classical methods based on RL rely on the capacity of the learning algorithm to extract mapping strategies from reward alone. While this has been shown to be possible in principle~\cite{DBLP:conf/pkdd/BeechingD0020}, we will show that \textbf{the emergence of a spatial mapping strategy is significantly boosted through auxiliary tasks}, which require the agent to continuously reason on the presence of targets w.r.t. to its viewpoint --- see Figure \ref{fig:aux_tasks}.

We introduce three auxiliary tasks, namely estimating if a target object has already been observed since the beginning of the episode, and if it is the case, the relative direction and the Euclidean distance to this object. If an object is visible in the current observation, it will be helpful for training the agent to recognize it (discover its existence and relevance to the task) and estimate its relative position. More importantly, if the target object was seen in the past, the auxiliary supervision will encourage the learning of representations of the environment, either implicitly or explicitly predicted by the agent, that are better spatially structured and populated with more relevant semantic information,
leading to an update of the neural memory of the agent. 

We propose the following contributions: 
(i) we show that the auxiliary tasks improve the performance of previous neural baselines by a large margin, which even allows to reach the performance of (incomparable) agents using ground-truth oracle maps as input;
(ii) we show the consistency of the gains over different inductive biases, i.e. different ways to structure neural networks, reaching from simple recurrent models to agents structured with projective geometry. 
This raises the question whether spatial inductive biases are required or whether spatial organization can be learned; 
(iii) the proposed method reaches SOTA performance on the \textit{Multi-ON} task, and corresponds to the winning entry of the \textit{CVPR 2021 Multi-ON challenge} \footnote{\scriptsize{\url{http://multion-challenge.cs.sfu.ca/2021.html}}}. The \textit{Test-Standard} leaderboard \footnote{\scriptsize{\url{https://eval.ai/web/challenges/challenge-page/805/leaderboard/2202}}}, an explanatory video
\footnote{\scriptsize{\url{https://youtu.be/rzHZNATBec8}}}
and code \footnote{\scriptsize{\url{https://github.com/PierreMarza/teaching_agents_how_to_map}}} are available.

\section{RELATED WORK}
\label{sec:related}

\myparagraph{Visual navigation} has been extensively studied in robotics~\cite{bonin2008visual, thrun2005probabilistic}. An agent is placed in an unknown environment and must solve a specified task involving reaching positions based on visual input, where \cite{bonin2008visual} distinguish map-based and map-less navigation. 
Recently, many navigation problems have been posed as goal-reaching tasks~\cite{DBLP:journals/corr/abs-1807-06757}. The nature of the goal, its regularities in the environment and how it is communicated to the agent have a significant impact on required reasoning capacities of the agent \cite{DBLP:conf/icpr/BeechingD0020}. In \textit{Pointgoal}~\cite{DBLP:journals/corr/abs-1807-06757}, an agent must reach a location specified as relative coordinates, while \textit{ObjectGoal}~\cite{DBLP:journals/corr/abs-1807-06757} requires the agent to find an object of a particular semantic category.
Recent literature \cite{DBLP:conf/icpr/BeechingD0020, DBLP:conf/nips/WaniPJCS20} introduced new navigation tasks with two important characteristics, \textit{(i)} their sequential nature, i.e.~an episode is composed of a sequence of goals to reach, and \textit{(ii)} the use of external objects as target objectives, i.e. the objects to find are not part of the scanned 3D scenes used as environments, but are for example randomly placed coloured cylinders as in~\cite{ DBLP:conf/nips/WaniPJCS20}.

\textit{Multi-Object Navigation (Multi-ON)}~\cite{DBLP:conf/nips/WaniPJCS20} is a task requiring to sequentially retrieve objects, but unlike the \textit{Ordered K-item} task  ~\cite{DBLP:conf/icpr/BeechingD0020}, the order is not fixed between episodes. A sequential task is interesting as it requires the agent to remember and to map potential objects it might have seen while exploring the environment, as reasoning on them might be required in a later stage. Moreover, using external objects as goals prevents the agent from leveraging knowledge about the environment layouts, thus focusing solely on memory. Exploration is another targeted capacity as objects are placed randomly within environments. For these reasons, our work thus focuses on the new challenging \textit{Multi-ON} task~\cite{DBLP:conf/nips/WaniPJCS20}.

\myparagraph{Learning-free navigation} A recurrent pattern in methods tackling visual navigation~\cite{bonin2008visual, thrun2005probabilistic} is modularity, with different computational entities solving a particular sub-part of the problem. A module might map the environment, another one localize the agent within this map, a third one performing planning. Low-level control is also often addressed by a specialized sub-module. Known examples are based on Simultaneous Localization and Mapping (SLAM)~\cite{bresson2017simultaneous}.

\myparagraph{Learning-based navigation}
The task of navigation can be framed as a learning problem, leveraging the abilities of deep networks to extract regularities from a large amount of training data. Formalisms range from Deep Reinforcement Learning (DRL) \cite{DBLP:conf/iclr/MirowskiPVSBBDG17, DBLP:conf/iclr/JaderbergMCSLSK17, DBLP:conf/icra/ZhuMKLGFF17} to (supervised) Imitation Learning~\cite{DBLP:conf/nips/DingFAP19}.
Our work focuses on improving the training strategy of autonomous agents trained with DRL by augmenting the reward-based supervision signal with auxiliary losses that are related to the downstream task.

Such agents can be reactive~\cite{DBLP:conf/icra/ZhuMKLGFF17}, but recent work tends to augment agents with memory, which is a key component, in particular in partially-observable environments~\cite{SDMIA15-Hausknecht, pmlr-v48-oh16}. It can take the form of recurrent units~\cite{cho-etal-2014-learning}, or become a dedicated part of the system.
In the context of navigation, memory can fulfill multiple roles: holding a latent map-like representation of the spatial properties of the environment, as well as general high-level information related to the task (``\textit{did I already see this object?}''). Common representations are metric~\cite{DBLP:conf/iclr/ParisottoS18, DBLP:conf/pkdd/BeechingD0020, Henriques_2018_CVPR}, or topological~\cite{beeching2020learning, Chaplot_2020_CVPR}. Other work reduces assumptions about the necessary structure of the environment representation by using Transformers~\cite{vaswani2017attention} as a memory mechanism on episodic data~\cite{Fang_2019_CVPR}.

In contrast to end-to-end training, other approaches decompose the agent into sub-modules~\cite{Chaplot2020Learning, Chaplot_2020_CVPR} trained simultaneously with supervised learning~\cite{Chaplot_2020_CVPR} or a combination of supervised, reinforcement and imitation learning~\cite{Chaplot2020Learning}.
Somewhat related to our work, in~\cite{Chaplot_2020_CVPR}, a dedicated semantic score prediction module is proposed, which estimates the direction towards a goal and is explicitly used to decide which previously unexplored ghost node to visit next inside a topological memory. In contrast, in our work we propose to predict spatial metrics such as relative direction as an auxiliary objective to shape the learnt representations, instead of explicitly using those predictions at inference time.

\myparagraph{Learning vs. learning-free}
The differences in navigation performance between SLAM-based and learning-based agents have been studied before~\cite{DBLP:journals/corr/abs-1901-10915, Savva_2019_ICCV}. Even though trained agents begin to perform better than classical methods in recent studies~\cite{Savva_2019_ICCV}, arguments regarding efficiency of SLAM-based methods still hold \cite{DBLP:journals/corr/abs-1901-10915, SadekICRA2022}. Frequently hybrid methods are suggested~\cite{Chaplot2020Learning, Chaplot_2020_CVPR}. In contrast, we explore the question, whether mapping strategies can emerge naturally in end-to-end training through additional pretext tasks.

\myparagraph{Auxiliary tasks} can be combined with any downstream objective to guide a learning model to extract more useful representations as proposed in \cite{DBLP:conf/iclr/MirowskiPVSBBDG17, DBLP:conf/iclr/JaderbergMCSLSK17} 
to improve, both, data efficiency and overall performance. \cite{DBLP:conf/iclr/MirowskiPVSBBDG17} predict loop closure and reconstruct depth observations; Lample et al.~\cite{lample2017playing} also augment the DRQN model~\cite{SDMIA15-Hausknecht} with predictions of game features in fps games.
A potential drawback is the need for privileged information, which, however, is readily available in simulated environments~\cite{Savva_2019_ICCV}.
This is also the case in our work, where we access information during training on explored areas, positions of objects and of the agent, which, of course, is also used for reward generation in classical RL.

In \cite{DBLP:conf/iclr/JaderbergMCSLSK17}, unsupervised objectives are introduced, such as pixel or action features and reward prediction. \cite{aux_pointgoal_custom} introduce self-supervised auxiliary tasks to speed up the training on \textit{PointGoal}. 
They augment the base agent from~\cite{wijmans2019dd} with an inverse dynamics estimator as in~\cite{pathak2017curiosity}, a temporal distance predictor, and an action-conditional contrastive module, which must differentiate between positives, i.e.~real observations that occur after the given sequence, and negatives, i.e.~observations sampled from other timesteps. \cite{ye2021auxiliary} introduce auxiliary tasks for \textit{ObjectGoal}, building on top of~\cite{aux_pointgoal_custom} and introduce the action distribution prediction and generalized inverse dynamics tasks and coverage prediction.

Our work belongs to the group of supervised auxiliary tasks, with an application to 3D complex and photo-realistic environments, which was not the case of most concurrent methods. We also specifically target the learning of mapping and spatial reasoning through additional supervision, which has not been the scope of previous approaches.

\section{LEARNING TO MAP}
\label{sec:method}

\noindent
We target the \textit{Multi-ON} task~\cite{DBLP:conf/nips/WaniPJCS20}, where an agent is required to reach a sequence of target objects, more precisely coloured cylinders, in a certain order, and which was used for a recent challenge organized in the context of the CVPR 2021 Embodied AI Workshop. 
Compared to much easier tasks like \textit{PointGoal} or (Single) \textit{Object Navigation}, \textit{Multi-ON} requires more difficult reasoning capacities, in particular mapping the position of an object once it has been seen. The following capacities are necessary to ensure optimal performance: (i) mapping the object, i.e. storing it in a suitable latent memory representation; (ii) retrieving this location on request and using it for navigation and planning, including deciding when to retrieve this information, i.e.~solving a correspondence problem between sub-goals and memory representation.

The agent deals with sequences of objects that are randomly placed in the environment. At each time step, it only knows  the class of the next target, which is updated when reached. The episode lasts until either the agent has found all objects in the correct order or the time limit is reached. 

\subsection{SOTA agents in Multi-ON} 
\label{sota_agents}
\begin{figure}
    \centering
    \includegraphics[width=0.45\textwidth]{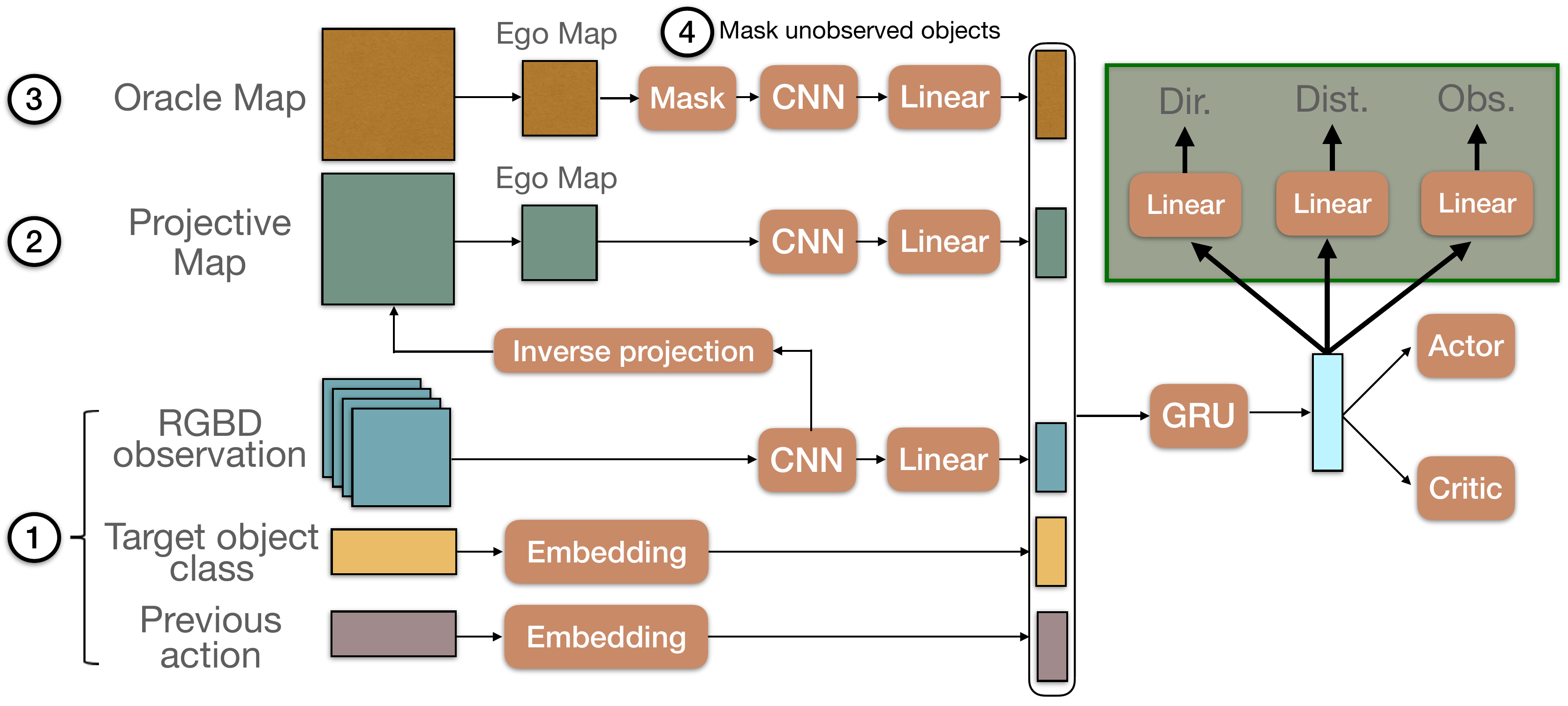}
    \caption{To study the impact of our auxiliary losses on different agents~\cite{DBLP:conf/nips/WaniPJCS20}, we explore several input and inductive biases. All variants share basic observations \ding{192} (RGB-D image, target class, previous action). Variants also use a map \ding{193} produced with inverse projective mapping.
    Oracle variants receive ground truth maps \ding{194}, where in one further variant unseen objects are removed \ding{195}. These architectures have been augmented with classification heads  implementing the proposed auxiliary tasks ({\color{RectangleGreen}\textbf{green rectangle}}).}
    \label{fig:architecture}
\end{figure}

\noindent
Our contribution is independent of the actual implementation choices in agents solving the \textit{Multi-ON} task as we rather target an improvement of the learning objective. We therefore explored several neural baselines with different architectures, as selected in~\cite{DBLP:conf/nips/WaniPJCS20}. The considered agents share a common base shown in Figure~\ref{fig:architecture}, which extracts information from the current RGB-D observation of the robot with a convolutional neural network (CNN) $f_o$, and computes embeddings of the target object class and the previous action taken by the agent. Differences between the considered baselines is in their representation of the environment. The simplest recurrent baseline \textit{NoMap} does not construct a map of its environment. \textit{OracleMap} and \textit{OracleEgoMap} baselines do not build a global map, but rather have access to oracle global maps of the environment containing channels for occupancy information and location of goal objects. Finally, \textit{ProjNeuralMap} builds a map of the environment in real time, associating feature vectors from $f_o$ with discrete cells in the spatial 2D representation using projective geometry. In variants that keep a global map, i.e. all except \textit{NoMap}, it is first transformed into an egocentric representation centered around the agent's position (explained further below). A vector representation of the map is then extracted using another CNN $f_m$. Such operation can be considered as a global read of the map.
The vector representations are concatenated and fed to a GRU~\cite{cho-etal-2014-learning} unit that integrates temporal information, and whose output serves as input to an actor and a critic heads, that respectively output a distribution over the set of actions to take and an estimation of the value of the state the agent is currently in. All agents are trained with the same RL algorithm (and same training hyper-parameter values) detailed in subsection~\ref{rl_training}, as well as the actor-critic formulation.

We present here in more details the considered variants which have been explored in~\cite{DBLP:conf/nips/WaniPJCS20}, but which have been introduced in prior work (numbers \ding{192}\ding{193}\ding{194}\ding{195} correspond to choices in Figure \ref{fig:architecture}):

\myparagraph{NoMap \ding{192}}  is a recurrent GRU baseline that does not explicitly build nor read a spatial map. The only memory available for storing mapping information is the flat vectorial hidden state of the GRU~\cite{cho-etal-2014-learning}, a variant of a recurrent neural network.
While the agent could in principle still learn (through RL) to use this vectorial memory like a spatial map, this is in no way enforced through any design choice.

\myparagraph{ProjNeuralMap \ding{192}\ding{193}~\cite{Henriques_2018_CVPR,DBLP:conf/pkdd/BeechingD0020}} is a neural network structured with spatial information and projective geometry. Or, stated in different terms, in this work the map is \textit{not} pre-computed by a handcrafted and engineered function (e.g. with estimated occupancy) and fed to an agent, as done in classical robotics; rather, the map is an internal activation of a neural network layer without trainable parameters. As such the content of the map is not predefined and interpretable through a handcrafted definition, the content is trained through machine learning, in our case RL. This layer is a map in the sense that (i) it is spatially organized and corresponds to an allocentric birds-eye representation, which is shifted and rotated with each agent motion through estimated odometry; (ii) using calibrated cameras, pixels are mapped to corresponding points on the map. However, the actual values stored at each position are determined through training and can, according to the learning signal, correspond to a latent representation of  anything ranging from occupancy to more semantic information like object positions. 

More specifically, ProjNeuralMap maintains a global allocentric map of the environment $M_{t} \in \mathbb{R}^{H{\times}W{\times}n}$ composed of $n$-channel vector representations, $n$ being an hyper-parameter, at each position within the full $H{\times}W$ environment.
Similar to Bayesian occupancy grids (BOG), which have been used in  mobile robotics for many years \cite{MoravecGrid1988},\cite{RummelhardCMCDOT2015}, the map is updated in the two-step process already mentioned above: (1) resampling taking into account estimated agent motion, and (2) integration of the representation of the current observation produced by a CNN $f_o$.

Writing to the map --- Given the current RGB observation $o_{t} \in \mathbb{R}^{h \times w \times 3}$, $f_o$ extracts an $n$-channel feature map $o^{\prime}_{t}$, which is then projected onto the 2D ground plane following the procedure in MapNet~\cite{Henriques_2018_CVPR} to obtain an egocentric map of the agent's spatial neighbourhood $m_{t} \in \mathbb{R}^{h^{\prime}{\times}w^{\prime}{\times}n}$.
The ground projection module assigns a discrete location on the ground plane to each element within $o^{\prime}_{t}$ conditioned on the input depth map $d_{t} \in \mathbb{R}^{h \times w}$ and known camera intrinsics. 
Registration of the observation $m_{t}$ to the global map is based on the assumption 
that the agent has access to odometry, as in~\cite{DBLP:conf/nips/WaniPJCS20}. The update to $M_{t}$ is performed through an element-wise max-pooling between $m_{t}$ and $M_{t-1}$.

Reading the map --- The global map is first cropped around the agent and oriented towards its current heading to form an egocentric map of its neighbourhood at time $t$, which is then fed to $f_m$ producing a context feature vector. The latter is then concatenated to the rest of the input, i.e. representations of the current observation, target object and previous action, producing the input to the recurrent memory (GRU) unit. The full model is trained end-to-end with Reinforcement Learning (RL), including networks involved in map writing and reading operations.

\myparagraph{OracleMap \ding{192}\ding{194}}  has access to a ground-truth grid map of the environment with $2$ channels. The first channel is dedicated to occupancy information with a binary value per cell indicating the presence of free space or an obstacle. The second channel encodes the presence of objects and their classes with thus $9$ possible values per cell, i.e. $1$ to $8$ for one of the $8$ object classes, or $0$ for no object. Each channel information is passed through a learned embedding layer to output a $m$-dim vector, $m$ being an hyperparameter, as it is common practice to represent categorical data fed to a neural network. This leads to a map with $2 \times m$ channels. The map is cropped and centered around the agent to produce an egocentric map as input to the model.

\myparagraph{OracleEgoMap \ding{192}\ding{194}\ding{195}}  gets the same egocentric map as OracleMap with only object channels, and revealed in regions that have already been within its field of view during the episode. This variant corresponds to an agent capable of perfect mapping --- no information gets lost, but only observed information is used.

Table~\ref{tab:summary_agents} summarizes the environment representation strategies used by the different baselines.

\begin{table}[t] \centering
\vspace*{3mm}
\caption{Summary of env. representation in SOTA baseline agents.}
{\small
\resizebox{0.7\textwidth}{!}{\begin{minipage}{\textwidth}
\begin{tabular}{ c c c c c c c c c }
\toprule
Agent & \multicolumn{1}{p{1cm}}{\centering \textbf{GRU \\ state}} & \textbf{Map} & \multicolumn{1}{p{1cm}}{\centering \textbf{Map \\ update}} & \multicolumn{1}{p{1cm}}{\centering \textbf{Map \\ reading}} & \multicolumn{1}{p{1.5cm}}{\centering \textbf{Full \\ visibility}} & \multicolumn{1}{p{1.5cm}}{\centering \textbf{Oracle \\ occupancy}} & \multicolumn{1}{p{1cm}}{\centering \textbf{Oracle \\ goals}} & \multicolumn{1}{p{1cm}}{\centering \textbf{Neural \\ features}} \\
\midrule

NoMap & ${\checkmark}$ & ${-}$ & ${-}$ & ${-}$ & ${-}$ & ${-}$ & ${-}$ & ${-}$ \\
OracleMap & ${\checkmark}$ & ${\checkmark}$ & ${-}$ & ${\checkmark}$ & ${\checkmark}$ & ${\checkmark}$ & ${\checkmark}$ & ${-}$ \\
OracleEgoMap & ${\checkmark}$ & ${\checkmark}$ & ${-}$ & ${\checkmark}$ & ${-}$ & ${-}$ & ${\checkmark}$ & ${-}$ \\
ProjNeuralMap & ${\checkmark}$ & ${\checkmark}$ & ${\checkmark}$ & ${\checkmark}$ & ${-}$ & ${-}$ & ${-}$ & ${\checkmark}$ \\

\bottomrule
\end{tabular}
\end{minipage}}

}
\label{tab:summary_agents}
\end{table}

\subsection{Learning to map objects with auxiliary tasks}
\noindent
We introduce auxiliary tasks, additional to the classical RL objectives, and formulated as classification problems, which require the agent to predict information on object appearances, which were in its observation history in the current episode. To this end, the base model is augmented with three classification heads (Figure~\ref{fig:architecture}) taking as input the contextual representation produced by the GRU unit. It is important to note that these additional classifiers are only used at training time to encourage the learning of spatial reasoning. At inference time, i.e. when deploying the agent on new episodes and/or environments, predictions about already seen targets, their relative direction are distance are not considered. Only the output of the actor is taken into account to select actions to execute.

\myparagraph{Direction}
the agent predicts the relative direction of the target object, only if it has been within its field of view in the observation history of the episode (Figure \ref{fig:aux_tasks} left). The ground-truth direction towards the goal is computed as,
\begin{equation}
\phi_t = \sphericalangle (\mathbf{o}_t,\mathbf{e}) = 
- \operatorname{atan2} (\mathbf{o}_{t, x} - \mathbf{e}_x, \mathbf{o}_{t, y} - \mathbf{e}_y) 
\end{equation} where $\mathbf{e} = [ \mathbf{e}_x \ \mathbf{e}_y ]$ (``\textit{ego}'')  are the coordinates of the agent on the grid and  $\mathbf{o} = [ \mathbf{o}_{t, x} \ \mathbf{o}_{t, y} ]$ are the coordinates of the center of the target object at time $t$. As the ground-truth grid is egocentric, the position of the agent is fixed, i.e.~at the center of the grid, while the target object gets different coordinates with time. The angles are kept in the interval $[0, 2\pi]$ and then discretized into $K$ bins, giving the angle class. The ground-truth one-hot vector is denoted $\phi_t^*$. At time instant $t$, the probability distribution over classes $\hat{\phi}_t$ is predicted from the GRU hidden state $\mathbf{h}_t$ through an MLP as $p(\hat{\phi}_t) = f_{\phi}(\mathbf{h_t}; \theta_{\phi})$ with parameters $\theta_{\phi}$.

\myparagraph{Distance}
The second task requires the prediction of the Euclidean distance in the egocentric map between the center box, i.e. position of the agent, and the mean of the grid boxes containing the target object (Figure \ref{fig:aux_tasks} right) that was observed during the episode,
$
d_t = || \mathbf{o}_t - \mathbf{e} ||_2
$.
Again, distances are discretized into $L$ bins, with $d_t^*$ as ground-truth one-hot vector, and at time instant $t$, the probability distribution over classes $\hat{d}_t$ is predicted from the hidden state $\mathbf{h}_t$ through an MLP as $p(\hat{d}_t) = f_{d}(\mathbf{h_t}; \theta_{d})$ with parameters $\theta_{d}$.

\myparagraph{Observed target}
This third loss favors learning whether the agent has previously encountered the target object. The model is required to predict the binary value $\mathds{1}_{t}^{\text{obs}}$, defined as $1$ if the target object at time $t$ has been within the agent's field of view at least once in the episode, and $0$ otherwise. The model predicts the probability distribution over classes $\hat{obs}_{t}$ given the hidden GRU state $\mathbf{h}_t$ through an MLP as $p(\hat{obs}_{t}) = f_{obs}(\mathbf{h_t}; \theta_{obs})$ with parameters $\theta_{obs}$.

\subsection{Training agents with Deep RL}
\label{rl_training}
\noindent
Following~\cite{DBLP:conf/nips/WaniPJCS20}, all agents are trained with Proximal Policy Optimization (PPO)~\cite{schulman2017proximal} and a reward composed of three terms,
\begin{equation}
R_{t}=\mathds{1}_{t}^{\text{reached}} \cdot R_{\text {goal }}+R_{\text {closer }}+R_{\text {time-penalty }}
\label{eq:eq_reward_ppo}
\end{equation}
where $\mathds{1}_{t}^{\text{reached}}$ is the indicator function whose value is $1$ if the \textit{found} action was called at time $t$ while being close enough to the target, and $0$ otherwise. $R_{\text {closer }}$ is a reward shaping term equal to the decrease in geodesic distance to the next goal compared to previous timestep. Finally, $R_{\text {time-penalty }}$ is a negative slack reward to force the agent to take short paths.

PPO alternates between sampling and optimization phases. At sampling time $k$, a set $\mathcal{U}_k$ of trajectories $\tau$ with length $T$ are collected using the latest policy $\pi_{\theta}$ where $\theta$ denotes the set of weights of the policy neural network. Note that $T$ is smaller than the length of a full episode. The base PPO loss is then,
\begin{equation}
  \mathcal{L}_{PPO} = \frac{1}{\left|\mathcal{U}_{k}\right| T} \sum_{\tau \in \mathcal{U}_{k}} \sum_{t=0}^{T-1}
\left[\min \left(r_{t}(\theta) \hat{A}_{t}, \mathcal{C}(r_{t}(\theta), \epsilon) \hat{A}_{t}\right)\right]
\end{equation} where $\mathcal{C}(r_{t}(\theta), \epsilon) = \operatorname{clip}\left(r_{t}(\theta), 1-\epsilon, 1+\epsilon\right)$, $\hat{A}_{t}$ is an estimate of the advantage function $A^{\pi_{\theta}}(s_t, a_t)=Q^{\pi_{\theta}}(s_t, a_t)-V^{\pi_{\theta}}(s_t)
$ at time $t$ with $Q^{\pi_{\theta}}(s_t, a_t) = \mathbb{E}_{a_{t'} \sim \pi_{\theta}}\left[\sum_{t'=t}^{T} \gamma^{t'} R_{t'} \mid S_{t}=s_t, A_{t}=a_t\right] $, $V^{\pi_{\theta}}(s_t)=\mathbb{E}_{a_{t'} \sim \pi_{\theta}}\left[\sum_{t'=t}^{T} \gamma^{t'} R_{t'} \mid S_{t}=s_t\right]$, and $r_{t}(\theta)=\frac{\pi_{\theta}\left(a_{t} \mid s_{t}\right)}{\pi_{\theta_{\text {old }}}\left(a_{t} \mid s_{t}\right)}$ is the probability ratio between the updated and old versions of the policy. $\gamma$ is referred to as the discount factor, $s_t$ and $a_t$ respectively denote the state and action at time $t$ within the trajectory. We did not make the dependency of states and actions on 
$\tau$ explicit in the notation.

We provide more details here regarding the actor and critic heads in the base architecture shared by all the considered agents. These two modules respectively predict a distribution $\pi_{\theta}(a_t \mid s_t)$ over actions $a_t$ conditioned on the current state $s_t$ and the state-value function $V^{\pi_{\theta}}(s_t)$, i.e.~expected cumulative reward starting in $s_t$ and following policy $\pi_{\theta}$. Combining an actor and a critic is a common approach in RL~\cite{sutton2018reinforcement}.

\subsection{Modification of the training objective with auxiliary tasks}
\noindent
We now detail our contribution, i.e. additional terms to the base PPO loss in order to encourage spatial reasoning in trained agents.

Direction, distance and observed target predictions are supervised with cross-entropy losses from ground truth values $\phi_t^*$, $d_t^*$ and $\mathds{1}_{t}^{\text{obs}}$, respectively, as
\begin{align}
\mathcal{L}_{\phi} &= \frac{1}{\left|\mathcal{U}_{k}\right| T} \sum_{\tau \in \mathcal{U}_{k}} \sum_{t=0}^{T-1}\left[ -
\mathds{1}_{t}^{\text{obs}} \sum_{c=1}^{K} \phi_{t,c}^* \log  p(\hat{\phi}_{t,c})\right] \\
\mathcal{L}_{d} &= \frac{1}{\left|\mathcal{U}_{k}\right| T} \sum_{\tau \in \mathcal{U}_{k}} \sum_{t=0}^{T-1}\left[ -
\mathds{1}_{t}^{\text{obs}} \sum_{c=1}^{L} d_{t,c}^* \log  p(\hat{d}_{t,c})\right] \\
\mathcal{L}_{obs} &= \frac{1}{\left|\mathcal{U}_{k}\right| T} \sum_{\tau \in \mathcal{U}_{k}} \sum_{t=0}^{T-1} - (\mathds{1}_{t}^{\text{obs}} \log p(\hat{obs}_{t}) +  \nonumber \\
    &\hspace{8em} (1-\mathds{1}_{t}^{\text{obs}}) \log (1-p(\hat{obs}_{t})))
\label{eq:eq_cross_entropy}
\end{align} where $\mathds{1}_{t}^{\text{obs}}$ is the binary indicator function specifying whether the current target object has already been seen in the current episode ($\mathds{1}_{t}^{\text{obs}}{=}1$), or not ($\mathds{1}_{t}^{\text{obs}}{=}0$). 

The auxiliary losses $\mathcal{L}_{\phi}$, $\mathcal{L}_{d}$ and $\mathcal{L}_{obs}$ are added as follows,
\begin{equation}
\mathcal{L}_{tot} = \mathcal{L}_{PPO} +
\lambda_{\phi}\mathcal{L}_{\phi} + \lambda_{d} \mathcal{L}_{d} + \lambda_{obs} \mathcal{L}_{obs}
\end{equation} where $\lambda_{\phi}$, $\lambda_{d}$ and $\lambda_{obs}$ weight the relative importance of auxiliary losses.

\section{EXPERIMENTAL RESULTS}
\label{sec:experiments}
\noindent
We focus on the \textit{3-ON} version of the \textit{Multi-ON} task, where the agent deals with sequences of $3$ objects. The time limit is fixed to $2500$ environment steps, and there are $8$ object classes.
The agent receives a $(256{\times}256{\times}4)$ RGB-D observation and the one-in-K encoded class of the current target object within the sequence. 
The discrete action space is composed of four actions:  \textit{move forward} $0.25m$, \textit{turn left} $30\degree$, \textit{turn right} $30\degree$, and \textit{found}, which signals that the agent considers the current target object to be reached.
As the aim of the task is to focus on evaluating the importance of mapping, a perfect localization of the agent was assumed as in the protocol proposed in~\cite{DBLP:conf/nips/WaniPJCS20}. 

\begin{figure*}
    \centering
    \includegraphics[width=\textwidth]{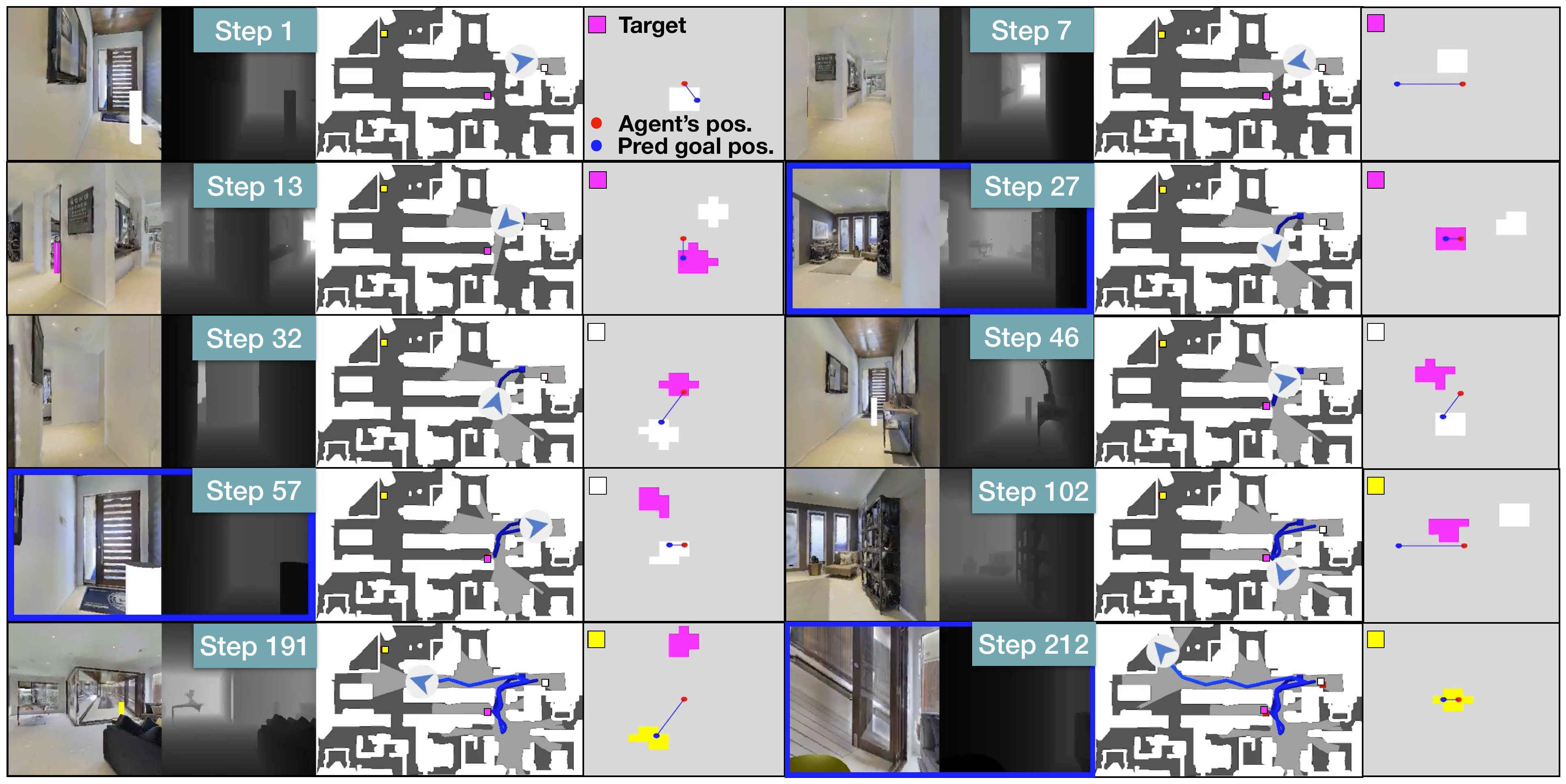}
    \caption{Example agent trajectory (sample from competition Mini-val set). The agent properly explores the environment to find the pink object. It then successfully backtracks to reach the white cylinder, and finally goes to the yellow one after another exploration phase (see text for a detailed description).  In columns 3 and 6, the relative direction and distance predictions are combined into a visualised blue point on top of the oracle egocentric map (Ground-truth object positions). The red point corresponds to the position of the agent. Note that these predictions are not used by the agent at inference time, and are only shown for visualisation purposes. The top down view and oracle egocentric map are also provided for visualisation only.}
    \label{fig:agent_examples}
\end{figure*}

\myparagraph{Dataset and metrics} we used the standard train/val/test split over scenes from the Matterport~\cite{chang2018matterport3d} dataset, ensuring no scene overlap between splits. There are  $61$ training scenes, $11$ validation scenes, and $18$ test scenes. The train split consists of $50,000$ episodes per scene, while there are $12,500$ episodes per scene in the val and test splits. Reported results on the val and test sets (Tables~\ref{tab:ablation} and \ref{tab:results}) were computed on a subset of $1,000$ randomly sampled episodes. Fig.~\ref{fig:agent_examples} shows an example of episode (from the Mini-val set of the \textit{CVPR 2021 Multi-On Challenge}) with RGB-D inputs.

We consider standard metrics of the field as  given in~\cite{DBLP:conf/nips/WaniPJCS20}:
\begin{itemize}[noitemsep]
    \item \textit{Success}: percentage of successful episodes (all three objects reached in the right order in the time limit).
    \item \textit{Progress}: percentage of objects successfully found in the right order in an episode.
    \item \textit{SPL}: Success weighted by Path Length. This extends the original SPL metrics from \cite{DBLP:journals/corr/abs-1807-06757}  to the sequential multi-object case.
     \item \textit{PPL}: Progress weighted By Path Length.
\end{itemize}
Note that for an object to be considered found, the agent must take the \textit{found} action while being within $1.5$m of the current goal. The episode ends immediately if the agent calls \textit{found} in an incorrect location. For more details, we refer to~\cite{DBLP:conf/nips/WaniPJCS20}.

\myparagraph{Implementation details}
training and evaluation hyper-parameters, as well as architecture details have been taken from~\cite{DBLP:conf/nips/WaniPJCS20}. All reported quantitative results are obtained after $4$ training runs ($6$ runs were computed for \textit{ProjNeuralMap} with the three auxiliary losses for job scheduling reasons) for each model, during $70M$ steps (increased from $40M$ in~\cite{DBLP:conf/nips/WaniPJCS20}). This amount of training time is standard when considering previous work targeting visual navigation with learning-based agents trained with RL.
Ground-truth direction and distance measures are respectively split into $K=12$ and $L=36$ classes. Indeed, angle bins span $30\degree$, and distance bins span a unit distance on the egocentric map, that is $50 \times 50$ (the maximum distance between center and a grid corner is thus $35$). The map used to compute ground-truth labels for auxiliary losses is the one fed to the \textit{OracleEgoMap} agent. Training weights $\lambda_{\phi}$, $\lambda_{d}$ and $\lambda_{obs}$ are all fixed to $0.25$. Each classification head is a single linear layer followed by a softmax activation function.

\begin{table}[t] \centering
\vspace*{3mm}
\caption{Impact of different auxiliary tasks (validation performance). 
The $\dagger$ column specifies comparable agents.}
{\small
\resizebox{0.78\textwidth}{!}{\begin{minipage}{\textwidth}
\begin{tabular}{ c c c c c c c c c }
\toprule
Agent & \textbf{Dir.} & \textbf{Dist.} & \textbf{Obs.} & \textbf{Success} & \textbf{Progress} & \textbf{SPL} & \textbf{PPL} & $\dagger$ \\
\midrule
OracleMap$^*$ & ${-}$ & ${-}$ & ${-}$ & $44.9 \pm$ {\footnotesize 1.7} & $55.7 \pm$ {\footnotesize 2.4} & $35.4 \pm$ {\footnotesize 1.4} & $43.7 \pm$ {\footnotesize 2.2} & ${-}$\\

OracleEgoMap$^*$ & ${-}$ & ${-}$ & ${-}$ &  $27.5 \pm$ {\footnotesize 2.7} & $42.8 \pm$ {\footnotesize 2.8} & $21.3 \pm$ {\footnotesize 2.5} & $32.7 \pm$ {\footnotesize 2.9} & ${-}$ \\
\thinrule

\multirow{4}{*}{ProjNeuralMap} & ${-}$ & ${-}$ & ${-}$ & $21.8 \pm$ {\footnotesize 1.7} & $38.6 \pm$ {\footnotesize 1.3} & $15.4 \pm$ {\footnotesize 0.7} & $27.0 \pm$ {\footnotesize 0.7} & ${\checkmark}$ \\

 & ${-}$ & ${-}$ & ${\checkmark}$ & $22.4 \pm$ {\footnotesize 2.9} & $40.2 \pm$ {\footnotesize 2.2} & $16.2 \pm$ {\footnotesize 2.7} & $28.9 \pm$ {\footnotesize 2.3} & ${\checkmark}$ \\
 
 & ${-}$ & ${\checkmark}$ & ${-}$ & $27.3 \pm$ {\footnotesize 3.3} & $43.0 \pm$ {\footnotesize 3.6} & $19.2 \pm$ {\footnotesize 2.1} & $30.6 \pm$ {\footnotesize 2.4} & ${\checkmark}$ \\

 & ${\checkmark}$ & ${-}$ & ${-}$ & $40.2 \pm$ {\footnotesize 4.2} & $55.9 \pm$ {\footnotesize 3.5} & $26.1 \pm$ {\footnotesize 2.2} & $36.4 \pm$ {\footnotesize 2.0} & ${\checkmark}$ \\
 
 & ${\checkmark}$& ${\checkmark}$ & ${-}$ & $44.3 \pm$ {\footnotesize 6.6} & $58.9 \pm$ {\footnotesize 4.9} & $29.0 \pm$ {\footnotesize 3.7} & $39.0 \pm$ {\footnotesize 2.2} & ${\checkmark}$ \\
 
 & ${\checkmark}$& ${\checkmark}$ & ${\checkmark}$ & \textbf{49.2 $\pm$ {\footnotesize 7.1}} & \textbf{62.8 $\pm$ {\footnotesize 5.2}} & \textbf{32.0 $\pm$ {\footnotesize 2.7}} & \textbf{41.1 $\pm$ {\footnotesize 1.1}} & ${\checkmark}$ \\
\bottomrule
\end{tabular}
\end{minipage}}
}
\label{tab:ablation}
\end{table}

\begin{table} \centering
\caption{Consistency over multiple models (test set).
The $\dagger$ column specifies comparable agents.}
{\small
\resizebox{0.83\textwidth}{!}{\begin{minipage}{\textwidth}
\begin{tabular}{ c c c c c c c }
\toprule
 Agent & \textbf{Aux. Sup.} & \textbf{Success} & \textbf{Progress} & \textbf{SPL} & \textbf{PPL} & $\dagger$ \\
\midrule
OracleMap$^*$ & ${-}$ & $50.4 \pm$ {\footnotesize 3.5} & $60.5 \pm$ {\footnotesize 3.1} & $40.7 \pm$ {\footnotesize 2.2} & $48.8 \pm$ {\footnotesize 1.9} & ${-}$ \\

 \thinrule
\multirow{2}{*}{OracleEgoMap$^*$} & ${-}$ & $32.8 \pm$ {\footnotesize 5.2} & $47.7 \pm$ {\footnotesize 5.2} & $26.1 \pm$ {\footnotesize 4.5} & $37.6 \pm$ {\footnotesize 4.7} & ${-}$ \\

 & ${\checkmark}$ & $44.0 \pm$ {\footnotesize 7.1} & $55.1 \pm$ {\footnotesize 7.0} & $35.0 \pm$ {\footnotesize 5.2} & $43.8 \pm$ {\footnotesize 5.0} & ${-}$ \\

 \thinrule

\multirow{2}{*}{ProjNeuralMap} & ${-}$ & $25.9 \pm$ {\footnotesize 1.1} & $43.4 \pm$ {\footnotesize 1.0} & $18.3 \pm$ {\footnotesize 0.6} & $30.9 \pm$ {\footnotesize 0.7} & ${\checkmark}$ \\

& ${\checkmark}$ & \textbf{57.7 $\pm$ {\footnotesize 3.7}} & \textbf{70.2 $\pm$ {\footnotesize  2.7}} & \textbf{37.5 $\pm$ {\footnotesize 2.0}} & \textbf{45.9 $\pm$ {\footnotesize 1.9}} & ${\checkmark}$ \\
 
 \thinrule
 
\multirow{2}{*}{NoMap} & ${-}$ & $16.7 \pm$ {\footnotesize 3.6} & $33.7 \pm$ {\footnotesize 3.3} & $13.1 \pm$ {\footnotesize 2.4} & $26.0 \pm$ {\footnotesize 1.7} & ${\checkmark}$ \\

 & ${\checkmark}$ & $43.0 \pm$ {\footnotesize 4.7} & $58.2 \pm$ {\footnotesize 4.0} & $29.5 \pm$ {\footnotesize 1.8} & $39.9 \pm$ {\footnotesize 1.3} & ${\checkmark}$ \\
\bottomrule
\end{tabular}
\end{minipage}}
}
\label{tab:results}
\end{table}

\begin{table}[t] \centering
\caption{\label{tab:results_challenge}CVPR 2021 \textit{Multi-ON} Challenge Leaderboard. \textit{Test Challenge} are the official challenge results. \textit{Test Standard} contains pre- and post-challenge results. Ranking is done with \textbf{PPL}.
The $^\ast$ symbol denotes Challenge baselines.}
{\small
\resizebox{0.9\textwidth}{!}{\begin{minipage}{\textwidth}
\setlength{\tabcolsep}{1pt}
\begin{tabular}{cccccccccc}
\toprule
Agent/Method & 
\multicolumn{4}{c}{--- Test Challenge ---} &&
\multicolumn{4}{c}{--- Test Standard ---} 
\\
&
Success & Progress & SPL & PPL  && 
Success & Progress & SPL & PPL  \\
\thinrule
Ours (Aux. losses) & 
\textbf{55}  & \textbf{67}  & \textbf{35}   & \textbf{44}  &&
57 & 70 & 36 & 45
\\
SGoLAM & 
52 & 64  & 32& 38 &&
62 & 71 & 34 & 39
\\
VIMP & 
41 & 57 & 26 & 36  &&
43 & 57 & 27 & 36
\\
ProjNeuralMap$^\ast$ &
${-}$ & ${-}$ & ${-}$ & ${-}$ &&
12 & 29 & 6  & 16
\\
NoMap$^\ast$ &
${-}$ & ${-}$ & ${-}$  & ${-}$ &&
5 & 19 & 3 & 13 
\\
\bottomrule
\end{tabular}
\end{minipage}}
}
\end{table}

\myparagraph{Do the auxiliary tasks improve the downstream objective?} in Table~\ref{tab:ablation}, we study the impact of the different auxiliary tasks on the 3-ON benchmark when added to the training objective of \textit{ProjNeuralMap}, and their complementarity. Direction prediction significantly improves performance, adding distance prediction further increases all metrics by a large margin, outperforming the performance of (incomparable) \textit{OracleEgoMap}. Both losses have thus a strong impact and are complementary, confirming the assumption that \textit{sense of direction} and \textit{judgement of relative distance} are two key skills for spatially navigating agents. The third loss about observed target objects brings a supplementary non-negligible boost in performance, showcasing the effectiveness of explicitly learning to remember, and its complementarity with distance and direction prediction.

Table~\ref{tab:results} presents results on the test set, confirming the significant impact on each of the considered metrics.\linebreak \textit{ProjNeuralMap} with auxiliary losses matches the performance of (incomparable) \textit{OracleMap} on Progress and Success, again outperforming \textit{OracleEgoMap} when considering all metrics.
\textit{OracleMap} has higher PPL and SPL, but has also access to very strong privileged information. 

Interestingly, \textit{OracleEgoMap} also benefits from the use of the auxiliary tasks at training time. As such agent already has access to priviledged information about the position of seen objects, this might suggest the auxiliary losses improve its spatial reasoning capabilities.

\myparagraph{Can an unstructured recurrent agent learn to map?} we explore whether an agent without spatial inductive bias, i.e. the assumption that the representation of the environment must be a 2D map, can be trained to learn a mapping strategy, to encode spatial properties of the environment into its unstructured hidden representation. As shown in Table~\ref{tab:results}, \textit{NoMap} indeed strongly benefits from the auxiliary supervision (Success for instance jumping from $16.7\%$ to $43.0\%$). Improvement is significant, outperforming \textit{ProjNeuralMap} trained without auxiliary supervision, and closing the gap with \textit{OracleEgoMap}. The quality of extra supervision can thus help to guide the learnt representation, mitigating the need for incorporating inductive biases into neural networks. When both are trained with our auxiliary losses, \textit{ProjNeuralMap} still outperforms \textit{NoMap}, indicating that spatial inductive bias still provides an edge. 

\myparagraph{Comparison with the state-of-the-art}
our method corresponds to the winning entry of the \textit{CVPR 2021 Multi-On Challenge} organized with the \textit{Embodied AI Workshop}, shown in Table~\ref{tab:results_challenge}. Test-standard is composed of $500$ episodes and Test-challenge of $1000$ episodes. In the context of the Challenge, the \textit{ProjNeuralMap} agent was trained for $80M$ steps with the auxiliary objectives, and then finetuned for $20M$ more steps with only the vanilla RL objective.
The official challenge ranking is done with \textbf{PPL}, which evaluates correct mapping (quicker and more direct finding of objects), while mapping does not necessarily have an impact on success rate, which can be obtained by pure exploration.

\myparagraph{Visualization} Figure~\ref{fig:agent_examples} illustrates an example trajectory from the agent trained with the auxiliary supervision in the context of the \textit{CVPR 2021 Multi-On Challenge}. The agent starts the episode (Step $1$) seeing the white object, which is not the first target to reach. It thus starts exploring the environment (Step $7$), until seeing the pink target object (Step $13$). Its prediction of the goal distance immediately improves, showing it is able to recognize the object within the RGB-D input. The agent then reaches the target (Step $27$). The new target is now the white object (that was seen in Step $1$). While it is still not within its current field of view, the agent can localize it quite precisely (Step $32$), and go towards the goal (Step $46$) to call the \textit{found} action (Step $57$). The agent must then explore again to find the last object (Step $102$). When the yellow cylinder is seen, the agent can estimate its relative position (Step $191$) before reaching it (Step $212$) and ending the episode.

\begin{figure}
    \centering
    \vspace*{3mm}
    \includegraphics[width=0.5\textwidth]{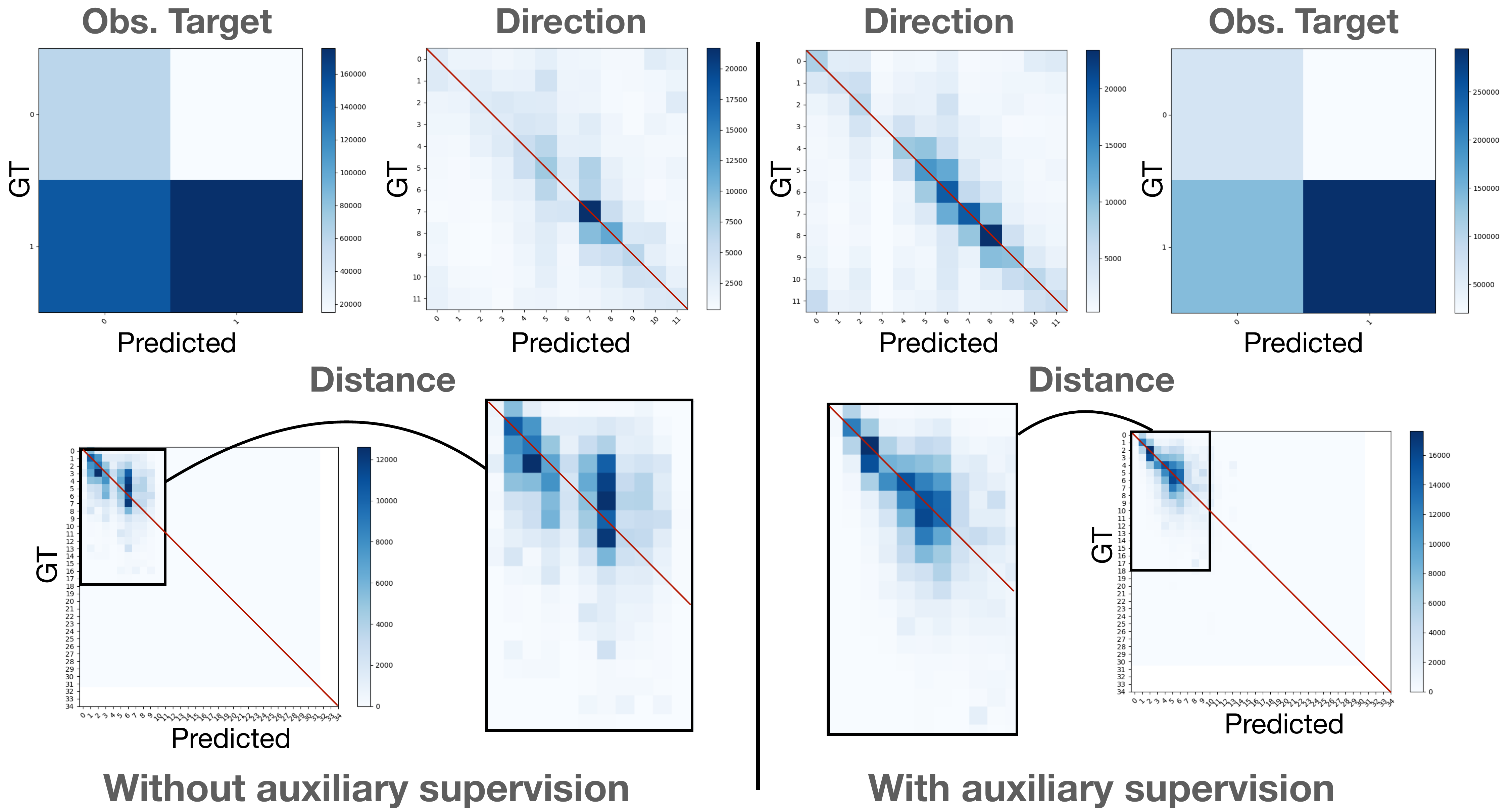}
    \caption{Confusion matrices (validation set) of linear probes trained on representations from both \textit{ProjNeuralMap} initially optimized with and without auxiliary supervision. Red lines indicate matrix diagonals.}
    \label{fig:probing}
\end{figure}

\myparagraph{Information about observed targets, their relative distance and direction} Is such knowledge extracted by \textit{ProjNeuralMap} without auxiliary supervision ? We perform a probing experiment by training three linear classifiers to predict this information from the contextual representation from the GRU unit, both for \textit{ProjNeuralMap} agent initially trained with and without auxiliary losses. We generate rollout trajectories on $1000$ training and validation episodes. 
It is important to note that, as both agents behave differently, linear probes are not trained and evaluated on the same data. Fig.~\ref{fig:probing} shows that linear probes trained on representations from our method perform better, and more consistently, suggesting the presence of more related spatial information.

\myparagraph{Last minute information} 
We discovered a bug in the official \textit{Multi-ON} code \cite{DBLP:conf/nips/WaniPJCS20} which in some cases provides too much information to the \textit{OracleEgoMap} baseline. This bug also affected the supervision of our agent (\textit{during training only, the bug maintains validity of agent}). The differences are small, do not change conclusions or method orders. New results for the values in Table \ref{tab:results} (test set) would be $52.3, 65.9, 36.4, 45.7$.

\section{CONCLUSION}
\label{sec:conclusion}
\noindent
In this work, we propose to guide the learning of mapping and spatial reasoning capabilities by augmenting vanilla RL training objectives with auxiliary tasks. We show that learning to predict the relative direction and distance of already seen target objects, as well as to keep track of those observed objects, improves significantly the performance on various metrics and that these gains are consistent over agents with or without spatial inductive bias. 
The proposed training strategy applied to a learning-based agent from the literature allowed us to win the \textit{CVPR 2021 Multi-ON challenge}.

\textbf{Acknowledgement} ---
We thank ANR for support through AI-chair grant ``Remember'' (\small{ANR-20-CHIA-0018}).

\bibliographystyle{IEEEtran}
\bibliography{ms}

\end{document}